\documentclass[letterpaper]{article} 
\usepackage{aaai23}  
\usepackage{times}  
\usepackage{helvet}  
\usepackage{courier}  
\usepackage[hyphens]{url}  
\usepackage{graphicx} 
\usepackage{natbib}  
\usepackage{caption} 
\usepackage{algorithm}

\usepackage[table]{xcolor}
\usepackage{epsfig}
\usepackage{graphicx}
\usepackage{amsmath}
\usepackage{amssymb}
\usepackage{bm}
\usepackage{algpseudocode}%
\usepackage{mathtools}
\usepackage{tabularx}
\usepackage{multirow}
\usepackage{booktabs}
\usepackage{multirow}

\usepackage{newfloat}
\usepackage{listings}

\DeclareCaptionStyle{ruled}{labelfont=normalfont,labelsep=colon,strut=off} 
\floatstyle{ruled}
\newfloat{listing}{tb}{lst}{}
\floatname{listing}{Listing}
\nocopyright

\title{\title{FaceCoresetNet: Differentiable Coresets for Face Set Recognition}}

\author {
    Gil Shapira\textsuperscript{\rm 1,2} and Yosi Keller \textsuperscript{\rm 1}
}
\affiliations {
    \textsuperscript{\rm 1} Bar-Ilan University, Ramat Gan, Israel\\
    \textsuperscript{\rm 2} Samsung Semiconductor Israel R\&D Center (SIRC)\\
    \{ggiillsshhaappiirraa, yosi.keller\}\@gmail.com
}

\usepackage{amsmath}
\usepackage{xcolor}
\usepackage{listings}

\makeatletter
\newcommand{\srcsize}{\@setfontsize{\srcsize}{8pt}{8pt}}
\makeatother
\lstdefinestyle{pytorchstyle}{
    language=Python,
    basicstyle=\ttfamily\srcsize,
    keywordstyle=\color{blue},
    commentstyle=\color{gray},
    stringstyle=\color{green},
    numbers=left,
    numberstyle=\scriptsize,
    stepnumber=1,
    numbersep=8pt,
    backgroundcolor=\color{white},
    frame=single,
    rulecolor=\color{lightgray},
    breaklines=true,
    breakatwhitespace=true,
    showspaces=false,
    showstringspaces=false,
    showtabs=false,
    tabsize=4
}

\begin{document}

\maketitle

\begin{abstract}
In set-based face recognition, we aim to compute the most discriminative descriptor from an unbounded set of images and videos showing a single person. A discriminative descriptor balances two policies when aggregating information from a given set. The first is a quality-based policy: emphasizing high-quality and down-weighting low-quality images. The second is a diversity-based policy: emphasizing unique images in the set and down-weighting multiple occurrences of similar images as found in video clips which can overwhelm the set representation.
This work frames face-set representation as a differentiable coreset selection problem. Our model learns how to select a small coreset of the input set that balances quality and diversity policies using a learned metric parameterized by the face quality, optimized end-to-end. The selection process is a differentiable farthest-point sampling (FPS) realized by approximating the non-differentiable Argmax operation with differentiable sampling from the Gumbel-Softmax distribution of distances. The small coreset is later used as queries in a self and cross-attention architecture to enrich the descriptor with information from the whole set. Our model is order-invariant and linear in the input set size.
We set a new SOTA to set face verification on the IJB-B and IJB-C datasets. Our code is publicly available \footnote{\url{https://github.com/ligaripash/FaceCoresetNet}}.
\end{abstract}

\begin{figure}
\includegraphics[width=.9\linewidth]{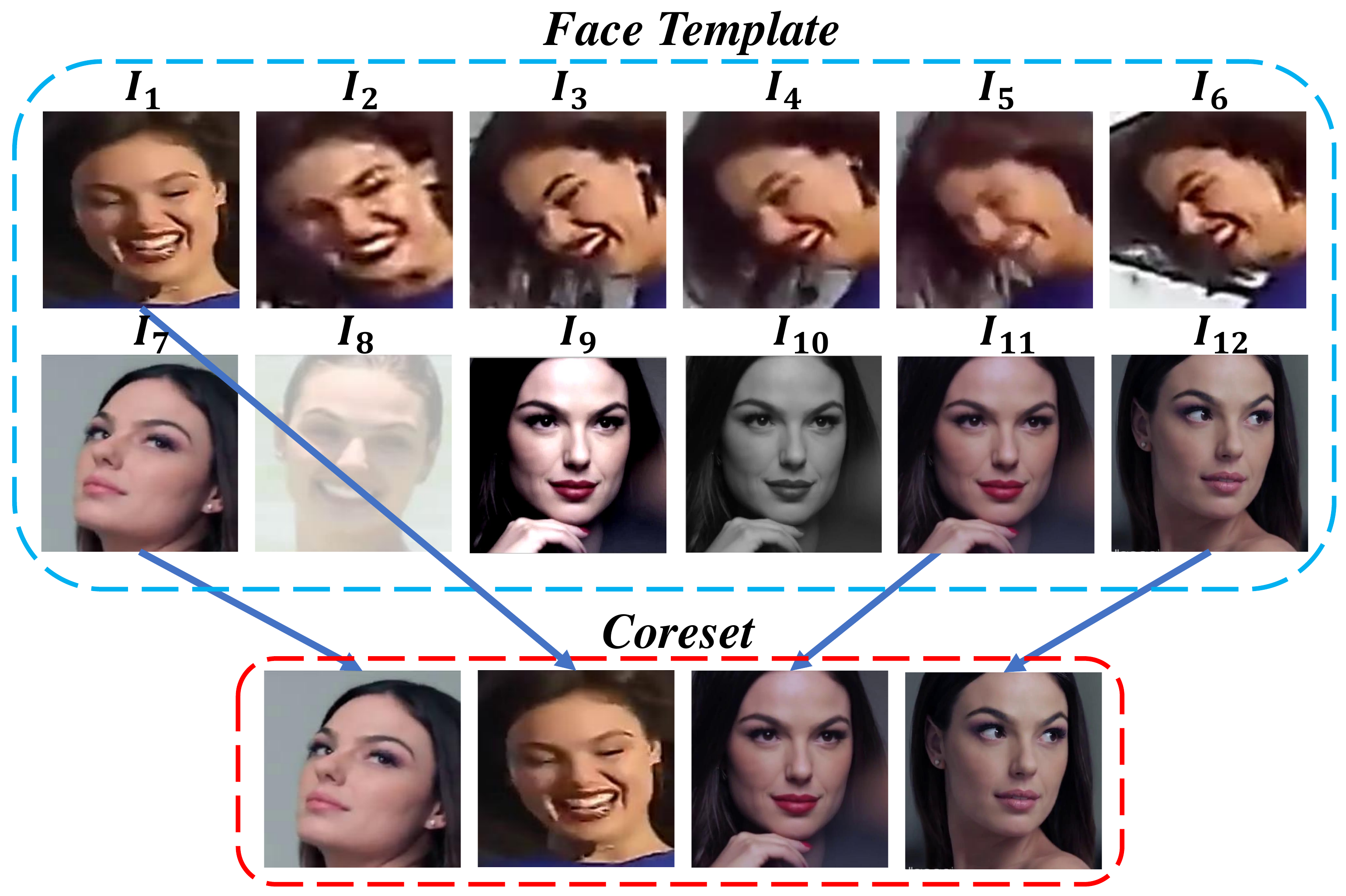}
\centering
\caption{
Template coreset selection. in the round blue rectangle, an input face set with 12 images containing a video clip sample ($I_1$ to $I_6$) and six photographs of the same person ($I_7$ to $I_{12}$). The coreset selection process balances quality and diversity. The final coreset in the red round rectangle contains four high-quality images with diverse poses and expressions.
}
\label{fig:figure_1}
\end{figure}


\section{Introduction}


Face Set Recognition (FSR) is a variant of Face Recognition (FR) that
involves recognizing an individual's identity based on a \textit{template},
that is an unbounded and unordered set of images and video clips depicting a
particular face instead of a single photograph. A template is shown in the Fig. %
\ref{fig:figure_1}. Given the added data, FSR is more accurate compared to
FR. The increased accuracy and proliferation of face media available on the
Internet and public cameras make FSR an attractive solution for 'in the
wild' situations, such as in unconstrained surveillance and Homeland Security
scenarios \cite{kalka2018ijb} where image quality is lacking due to low
resolution, extreme pose, illumination, occlusions, and recognition of a single image is difficult.

FSR consists of verification, which means confirming the identity of an
input query template (also referred to as a probe) and identification,
which entails searching a database of registered subjects (known as a
gallery) to identify an unknown probe person. FSR schemes encode the images in a template
into a fixed-size vector descriptor to allow computationally efficient similarity search. As the templates consist of an
arbitrary number of images and video frames, the image aggregation process
has to be efficient with respect to the template's size. Furthermore, since
the template is an unordered set of images, the aggregation should be
permutation invariant. Additionally, the aggregation should prioritize
informative images while downplaying uninformative, low-quality (blurry,
poorly lighted) images. Finally, the process of feature fusion should factor in "face burstiness". If a template includes a burst of similar frames as in video clips, the burst should not overpower the template descriptor.

Some aggregation schemes apply average pooling \cite{chen2015end_avgpool,
parkhi2015deep_avgpool} to the set of features. However, this method
presents a significant drawback. All images within the set are weighted
equally, regardless of their quality. To address this issue, \cite%
{kim2022adaface, meng2021magface} suggest facial features whose magnitude is
proportional to image quality. To improve the aggregation, other works model
intra-set relationships using reinforcement learning \cite{liu2018dependency}%
, recurrent models \cite{gong2019low}, or self-attention \cite%
{gong2019video, liu2019permutation, liu2017quality, xie2018multicolumn}. In contrast to intra-set modeling, face burstiness received only limited attention in the literature \cite{wang2022set}.

While Recurrent methods \cite{graves2012long, gong2019low} are effective in sequence modeling, they are unsuitable for modeling orderless sets as they are not permutation invariant. On the other hand, Multihead Self Attention (MSA) \cite{vaswani2017attention, liu2018dependency} is an effective permutation-invariant approach to model self-attention among template features, but it becomes impractical for large templates since it is quadratic in the template size.

In this work, we propose a feature aggregation approach, denoted \textit{FaceCoresetNet}, that selects a small coreset we dub \textit{Core-Template} of the face template with balanced quality and diversity. Given $N$ data points, a coreset is a smaller subset of $K<<N$ sampled points that approximate the original dataset with respect to a given metric. It is often used to speed up training or reduce the memory and storage requirements of large datasets \cite{mirzasoleiman2020coresets, campbell2018bayesian_coreset, pooladzandi2022adaptive_coreset}. 

We select the Core-Template using a greedy differentiable farthest point sampling (FPS) process with a learned metric parameterized by the quality of the face images. As computing the farthest point requires an Argmax operation, which is not differentiable, we suggest a novel differential FPS by replacing the Argmax operation with sampling from the differentiable Gumbel-Softmax \cite{jang2016categorical} distribution of quality-aware distances. Unlike conventional FPS which starts with a point selected at random, our selection process starts with highest quality feature, hence the selection is permutation invariant. An example of a face template and the selected Core-Template is depicted in Fig. \ref{fig:figure_1}.
As far as we know, this is the first differential coreset approach that can be fully optimized by backpropagation.
Importantly, our differentiable Core-Template selection reduces the unbounded template size $N$ to a fixed size $K$, facilitating the overall linear algorithm complexity.

To enhance the Core-Template representation, we compute self-attention amongst the Core-Template features. To enrich the representation with information from the full template missing from the Core-Template, we compute cross-attention between the Core-Template and the full template, using the Core-Template features as queries and the full template as Key-Values. By using the Core-Template features as queries for self-attention and cross-attention, we compute attention in linear time, unlike the typical quadratic time complexity in other works modeling intra-template relations, such as in the well-written paper by \cite{kim2022cluster_aggregate}.

As a final step, we sum and normalize the enriched Core-Template features to get the final feature representation for the template. The final template representative feature is used to compute the vanilla FR adaptive margin loss \cite{kim2022adaface} to optimize the model, which is simple and effective.
The Core-Template selection process and attention computation are linear in the template size, facilitating an overall linear time complexity of FaceCoresetNet.
The proposed scheme is shown to achieve SOTA accuracy on
the IJB-B \cite{ijb-b}, IJB-C \cite{ijb-c} FSR benchmarks.


In particular, we propose the following contributions:

\begin{itemize}

\item We propose a novel differential, permutation invariant, coreset selection for SFR, with a differentiable FPS strategy based on a quality-aware metric to balance quality and diversity. The coreset selection and metric parameters are part of the trained model and optimized end-to-end. As far as we know, this is the first differential coreset selection method with potential value to other fields. 
\item We model intra-template relations using the selected coreset as queries in self and cross-attention architecture to extract information from the whole set. 
\item As far as we know, our algorithm is the first to compute intra-template relations in linear time.
\item We use a vanilla FR AdaFace loss with no additional loss terms for a simple design.
\item We establish a new SoTA for face verification on IJB-B and IJB-C datasets

\end{itemize}

\begin{figure*}
\includegraphics[clip, trim=2cm 0.5cm 2cm 0.5cm,width=0.9\linewidth]{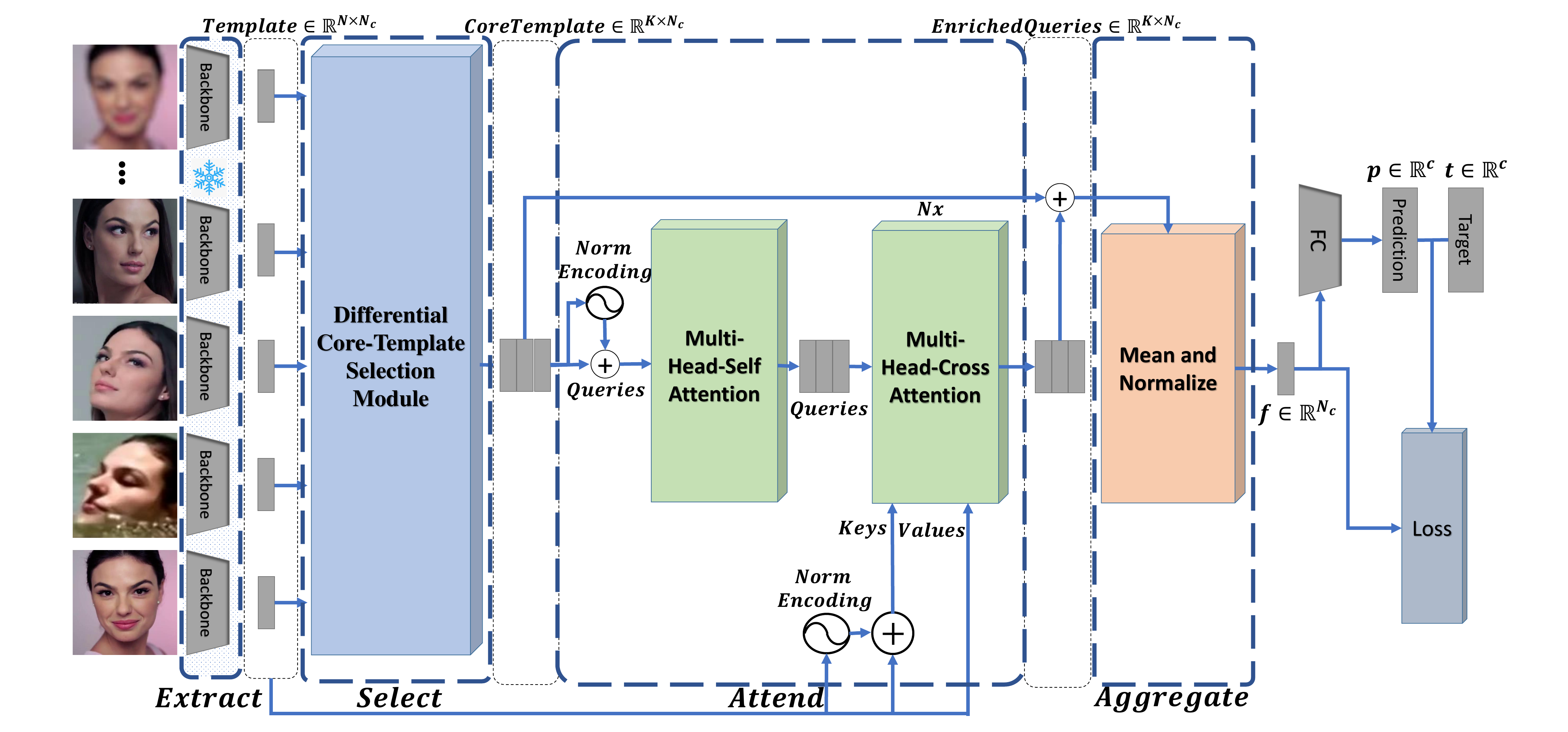}
\centering
\caption{FaceCoresetNet Architecture}
\label{fig:architecture}
\end{figure*}

\section{Related Work}

A simple method to combine features involves computing the mean of a collection of features \cite{chen2015end_avgpool,parkhi2015deep_avgpool}. A limitation of this elementary technique is that inferior-quality images can dominate the combined descriptor, diminishing its discriminatory capacity. During training with margin-based softmax, superior-quality faces indirectly produce features with greater magnitude \cite{meng2021magface, kim2022adaface}. By summing the features, high-quality images are inherently emphasized without needing explicit image quality evaluation. Both CFAN \cite{deng2019arcface} and QAN \cite{liu2017quality} employ a self-attention mechanism to learn image weighting through an explicit quality assessment \cite{gong2019video,liu2017quality}. Nevertheless, these approaches share a common drawback: they do not consider the relationships within the set during the weight computation stage.

To model intra-set relationships, the nonlocal neural network and RSA \cite{liu2019permutation, wang2018non} utilize intermediate feature maps $U_i$ of size ${C_m\times H\times W}$, during aggregation. This approach leverages the rich and complementary information feature maps provide and considers spatial relationships to refine them. However, the main limitation is the intensive computation required for attention calculation. Specifically, for a set of N feature maps, an attention module requires an affinity map of size ${(N\times H\times W)}^2$. 
CAFace, cluster and Aggregate network \cite{kim2022cluster_aggregate} generates an affinity map of size $N^2$, resulting in improved computation efficiency for attention calculation. Using our differential selection of coresets, this work reduces the feature fusion complexity to $O(N)$.


Set burstiness suppression network \cite{wang2022set}, tackles the face burstiness, a problem mostly overlooked by FSR research, by quantizing the face set feature Gram matrix, which has a size of $N^2$. In the main FSR datasets IJB-B, IJB-C, each image in each template is labeled by its media source (either a video clip or a singular image). MagFace and AdaFace \cite{kim2022adaface, meng2021magface} workaround the face burstiness problem by using these labels to fuse each media source separately and then sum the per-media features to produce the final fused feature. In this method, each media source contributes equally to the final descriptor, regardless of size, mitigating the burstiness problem.
In our work, we solve face burstiness in linear time without resorting to media labels, which may not be available, and still achieve superior performance.

DAC \cite{liu2018dependency} proposes an RL-based quality estimator, while MARN \cite{gong2019low} is an RNN-based quality estimator. However, these methods are not agnostic to input order, making them unsuitable for modeling orderless sets. In contrast, our method is permutation invariant, and well suited for set modeling.

A long-standing practice in fundamental approaches such as kNN and kMeans \cite{har2005smaller} or mixture models \cite{feldman2011scalable_coreset} is the use of subsets of the core to approximate the structure of an available set. These subsets enable the finding of approximate solutions with considerably reduced costs \cite{agarwal2005geometric}. Recently, coreset-based methods have made their way into Deep Learning techniques, such as network pruning \cite{mussay2021datacoreset}, active learning \cite{kim2022defense_coreset}, and increasing effective data coverage of mini-batches for improved GAN training \cite{sinha2020small_coreset}, or representation learning \cite{roth2020revisiting_coreset}. The latter two have achieved success through a greedy coreset selection mechanism. 

As far as we know, all previous coreset selection methods are non-differentiable. In contrast, in our proposed method, the coreset selection is integrated into our PyTorch model and trained end-to-end. Quality-based distance metric parameters are optimized during training, allowing an optimal balance between quality and diversity in the selected coreset.


To implement our differentiable Core-Template selection, we use the Gumbel-Softmax \cite{jang2016categorical} technique, which enables the smoothing of discrete categorical distributions, making them suitable for backpropagation. It employs the Gumbel-Max reparametrization trick, which efficiently draws samples from a Categorical distribution. Let $z$ be the categorical variable with class probabilities $\pi_1, \pi_2\dots\pi_k$. The trick asserts that sampling from the categorical distribution is equivalent to sampling from the following expression: 

$z = one\_hot(\underset{i}{\mathrm{argmax}}[g_i + log\pi_i])$

Here, $log\pi_i$ denotes the logits of the categorical distribution, and $g_i$ follows a Gumbel distribution with parameters (0,1). To avoid non-differentiability introduced by the Argmax operation, it is replaced with a Softmax function:

$y_i = \dfrac{exp((log(\pi_i) + g_i) / \tau)}{\Sigma_{j=1}^{k}exp((log(\pi_j) + g_j) / \tau)}$
for $i = 1,...,k$.

The temperature parameter ($\tau$) controls the spreading of the distribution. As $\tau$ approaches zero, the Gumbel-Softmax method converges to the discrete categorical distribution. During training, a $\tau$ value of 1 is typically used, while during inference, $\tau$ approaches zero ($\tau\rightarrow0^{+}$).
\cite{yang2019modeling} employ the Gumbel-Softmax for a differentiable sampling of point clouds, facilitating tasks such as point cloud segmentation and classification.

Multi-head self-attention (MSA) \cite{vaswani2017attention} is a commonly used function representing intra-set and inter-set relationships through an affinity map. It is a critical component of transformer architectures, which have surpassed CNNs in various visual tasks \cite{dosovitskiy2020vit, zhang2022dino, zheng2021rethinking}. In principle, each MSA output value is a weighted average of other values controlled by the queries and keys affinity. The MSA time complexity is $O(N^2)$; hence computing intra-set attention on a large template is prohibitive. To reduce the computation time, we compute intra-set attention on a fixed-size coreset, and cross-attention between the coreset and the template in $O(N)$ time.

\section{Method}

Let $\bm{T} = \{\bm{x}_1, \bm{x}_2, \ldots, \bm{x}_N\}$ be a template of $N$ facial images of the same identity. The goal is to produce a discriminative single compact feature vector $\bm{f}$ from $\bm{T}$. To compute $\bm{f}$, we employ an \textit{Extract, Select, Attend and Aggregate} paradigm.
To \textit{Extract} (fig. \ref{fig:architecture} left) compact feature $\bm{f_i}$ from each image $\bm{x}_i$, we use a pre-trained frozen single image FR model $E: \bm{x}_i \rightarrow \bm{f}_i$ following \cite{meng2021magface, kim2022adaface}. The set of features produced is $\bm{F} = \{\bm{f}_1, \bm{f}_2, \ldots, \bm{f}_N\}$ or in a matrix notation $\bm{F} \in \mathbb{R}^{N\times N_c}$ where $N_c$ is the number of channels in the compact feature. During training, we create a batch of $B$ templates of $B$ different individuals. Each template of size $N$ where $N_{min} \leq N \leq N_{max}$ , i.e. $\bm{F} \in \mathbb{R}^{B\times N\times N_c}$. The template size $N$ is randomly selected for each mini-batch.

\subsection{Core Template Selection} \label{subsection:core_template_selection}

To reduce the template feature size $\bm{F}$ from unbounded size $N$, we \textit{Select} (Fig. \ref{fig:architecture} green block) a core template denoted by $\bm{CT}$ of fixed size $K$. Our core template selection process strives to select high-quality images with diverse appearances to address the burstiness problem. As established in \cite{kim2022adaface}, $||\bm{f}_i||$ is a proxy for $\bm{x}_i$'s image quality. Our core template selection starts by selecting the feature $\bm{f}_i$ with the largest norm in $\bm{F}$, the template's highest quality feature. Following the farthest point sampling \cite{eldar1997farthest} paradigm, in each selection iteration, we choose the farthest feature in $\bm{F} \setminus \bm{CT}$ from the set of features already selected in $\bm{CT}$. The farthest point in our setting should correlate to a high-quality image with a diverse appearance relative to the features already selected for the core template. Cosine distance between two features $\bm{f}_i$ and $\bm{f}_j$ is defined as:
\begin{equation}
\begin{aligned}
  d_c(\bm{f}_i, \bm{f}_j) \coloneqq 1 - \frac{\bm{f}_i \cdot \bm{f}_j}{|| \bm{f}_i||||\bm{f}_j||}
\end{aligned}
\end{equation}

Large cosine distance between features correlates with large appearance variations (face pose, expression, age, etc.) between the corresponding images. Selecting the farthest feature using this metric encourages core template appearance diversification, but ignores image quality, resulting in the selection of diverse but low-quality images. 
To account for image quality, we scale the cosine distance by the feature norm exponentiated by a learned parameter $\gamma$ to balance between quality and diversity:

\begin{equation}
\label{eq:quality_aware_distance}
\begin{aligned}
  d_q(\bm{f}_i, \bm{f}_j;\gamma)  \coloneqq d_c(\bm{f}_i, \bm{f}_j)||\bm{f}_j||^{\gamma}
\end{aligned}
\end{equation}

\begin{figure}
\includegraphics[clip, trim=2cm 40cm 45cm 2cm,width=1.0\linewidth]{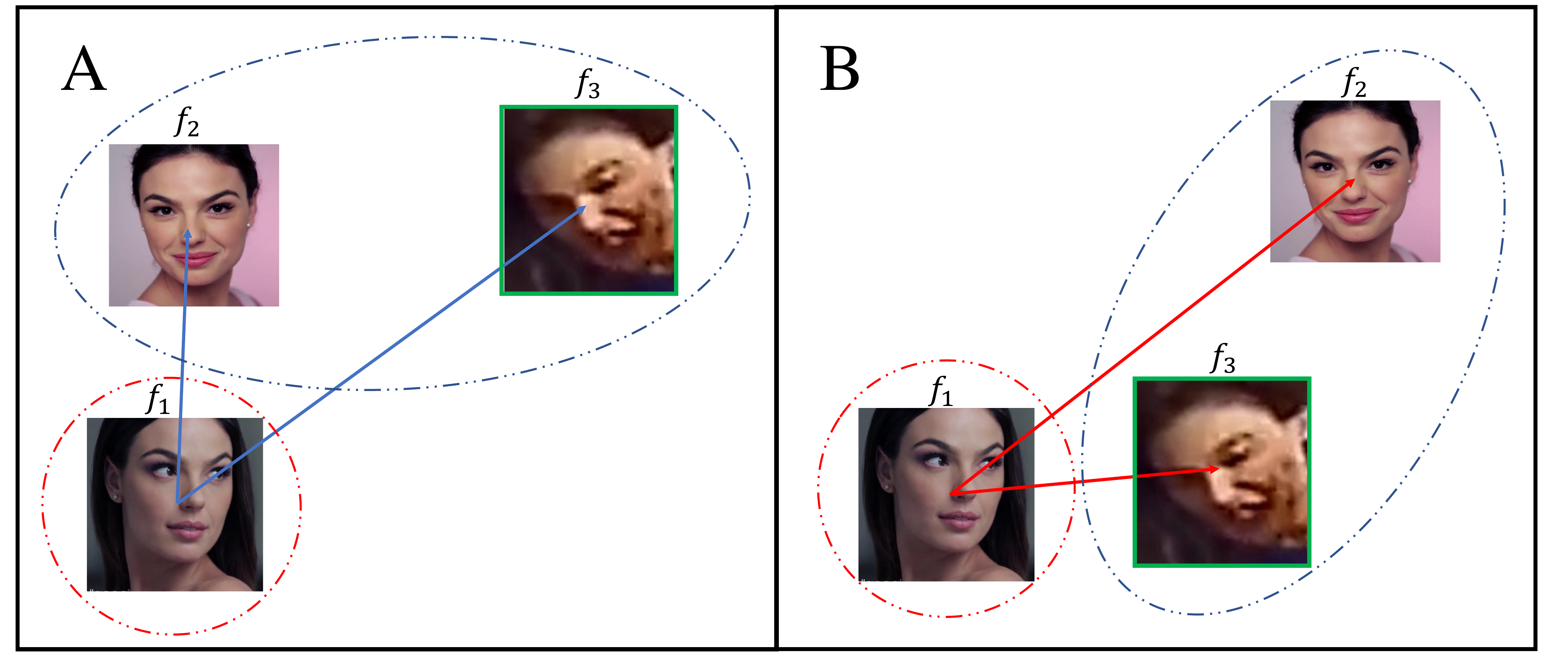}
\centering
\caption{Cosine distance (A) vs. quality-aware distance (B) for core template selection. The distance from $f_1$ to $f_2$ increases in B relative to A, as $f_2$ has high quality. The distance from $f_1$ to $f_3$ is decreased in B, as $f_3$ has low quality.}
\ref{subsection:core_template_selection}
\label{fig:figure_quality_aware_metric}
\end{figure}

\begin{figure}
\includegraphics[clip, trim=1cm 14.6cm 21cm 1cm,width=1.0\linewidth]{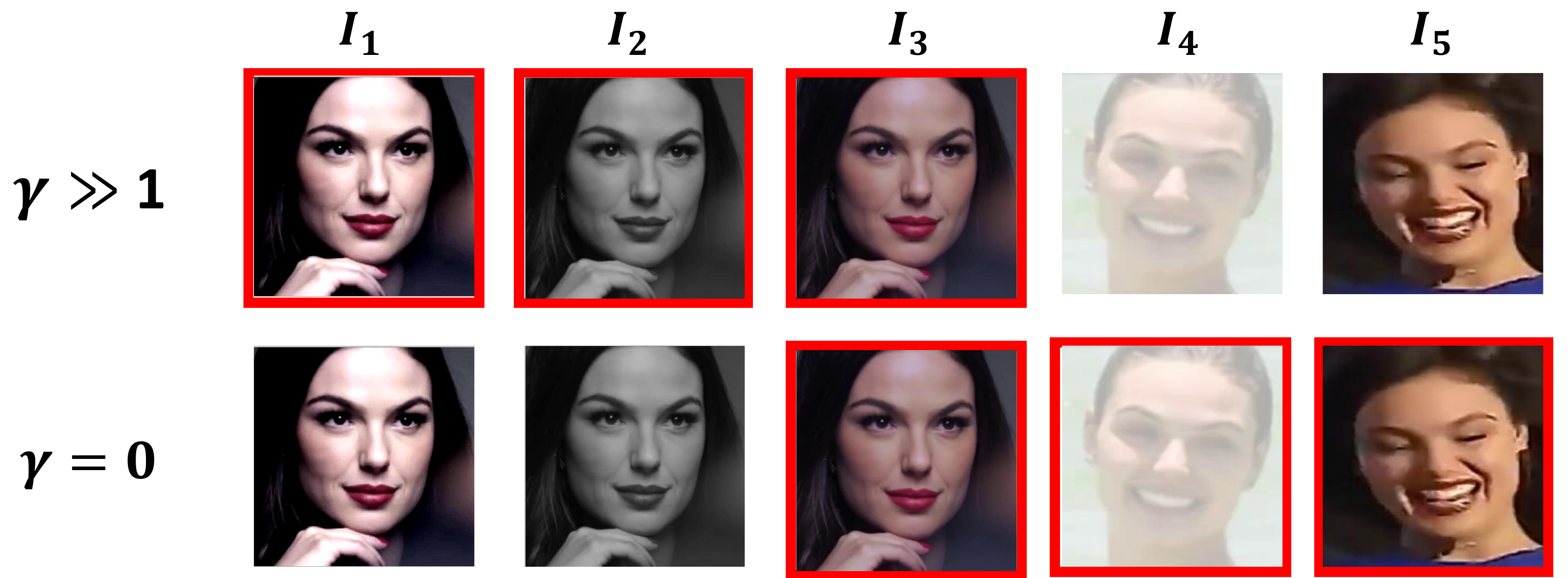}
\centering
\caption{The core template selection is influenced by the value of $\gamma$. In this example, the template size $N=5$ and $K=3$. In the top row, when $\gamma$ is large, the selection process prioritizes image quality over diversity. As a result, three similar high-quality images are chosen, marked with red rectangles. Conversely, in the bottom row, with $\gamma = 0$, diversity precedes image quality, leading to a selection that emphasizes a diverse range of images.}
\label{fig:selection_by_gamma}
\end{figure}

Here, $\bm{f}_i \in \bm{CT}$ and $\bm{f}_j \in \bm{F} \setminus \bm{CT}$ and $d_q$ is a non-symmetric distance function modulated by image $\bm{x}_j$'s quality. $\gamma$ is a trained parameter that reflects the optimal importance of the image quality. If $\gamma$ is 0, there is no weight to quality and $d_q$ is reduced to $d_c$. If $\gamma >> 1$, the feature quality dominates, and we have $d_q(\bm{f}_i, \bm{f}_j;\gamma) \approx ||\bm{f}_j||^{\gamma}$ as illustrated in Fig. \ref{fig:selection_by_gamma}.
The effect of using our quality aware distance function vs. regular cosine distance is illustrated in Fig. \ref{fig:figure_quality_aware_metric}. In Fig. \ref{fig:figure_quality_aware_metric}A, we have the template $\bm{F}=\{\bm{f}_1, \bm{f}_2, \bm{f}_3\}$ and $\bm{f}_1$ already in the Core Template ($\bm{CT} = \{\bm{f}_1\}$), marked by the red dashed ellipse, and we want to choose the next feature for the core template out of $\bm{F}\setminus \{\bm{f}_1\} = \{\bm{f}_2, \bm{f}_3\}$ (marked by the blue dashed ellipse). Using the regular cosine distance, the feature $\bm{f}_3$ of low-quality image 3 is farther from $\bm{f}_1$ than high-quality image 2 due to its larger pose difference. As a result, the low quality $\bm{f}_3$ feature is selected for CT ($\bm{CT} = \{\bm{f}_1, \bm{f}_3\}$), reducing the discriminative power of $\bm{CT}$. In Fig. \ref{fig:figure_quality_aware_metric}B, using our quality aware metric Eq. \ref{eq:quality_aware_distance}, the distance $d_c(\bm{f}_1, \bm{f}_2)$ is increased by $\bm{f}_2$'s larger quality measure, and the distance $d_c(\bm{f}_1, \bm{f}_3)$ is reduced by $\bm{f}_3$ lower quality, reversing the distance order and selecting the high quality $\bm{f}_2$ for the core template ($\bm{CT} = \{\bm{f}_1, \bm{f}_2\}$)

The Core-Template selection algorithm extends the FPS  algorithm with important modifications. Instead of relying on the non-differentiable Argmax operation to choose the feature in $\bm{T}$ farthest from $\bm{CT}$, we adopt a differentiable approach by sampling from the Gumbel-Softmax distribution of distances \cite{jang2016categorical}. This enables us to incorporate differentiability and leverage backpropagation during training.
Additionally, we introduce a quality-based distance function that incorporates a learned parameter $\gamma$, allowing the algorithm to achieve an optimal balance between quality and diversity in the selected Core-Template features.
Furthermore, unlike the conventional FPS method that randomly selects the first feature, we prioritize the feature with the highest norm (highest quality) to be inserted first into the Core-Template $\bm{CT}$. This selection criterion ensures permutation invariance and enhances the algorithm's overall performance.
The pseudocode for our core template selection algorithm 
is in Algorithm \ref{alg:core_template_selection}.


\begin{lstlisting}[style=pytorchstyle, caption={Differential Core-Template Selection}, label=alg:core_template_selection]

def Core-Template_selection(F, K, F_norms):
    """
    Input:
        F: normalized features data, [B, N, C]
        K: The intended coreTemaplate (CT) size (scalar)
        F_norms: F norms, [B, N]
        gamma: learned quality vs. diversity balance parameter (scalar)
    Return:
        CoreTemplate CT with size K
    """
    temperture = 1.0 if train else 1e-10
    #Soft select the feature with highest quality (max norm)
    1_hot_max = torch.gumbel_softmax(F_norms, tau=temperture, hard=True)        
    #Extract the feature with maximum norm
    CT = F @ 1_hot_max
    #After extraction compute the distance from CT to F
    d_CT_to_F = quality_aware_dist(CT, F_norms, F)
    for i in range(K-1):
        1_hot_max = torch.gumbel_softmax( d_CT_to_F, tau=temperture, hard=True)
        # extract next feature to add to CT
        new_f = F @ 1_hot_max
        d_new_f_to_CT = quality_aware_dist(new_f, norms, F, gamma)
        d_CT_to_F = torch.min(d_CT_to_F, d_new_f_to_CT)
        CT = torch.cat([CT, new_f])
    return CT
    
def quality_aware_dist(candidate_point, F_norms, F, gamma):
    inner_product = torch.bmm(F, candidate_point)
    cosine_dist = (1 - inner_product)
    quality_dist = torch.pow(norms, gamma) * cosine_dist
    return quality_dist
\end{lstlisting}

\begin{figure}
\includegraphics[clip, trim=0.5cm 5cm .5cm 1cm,width=0.8\linewidth]{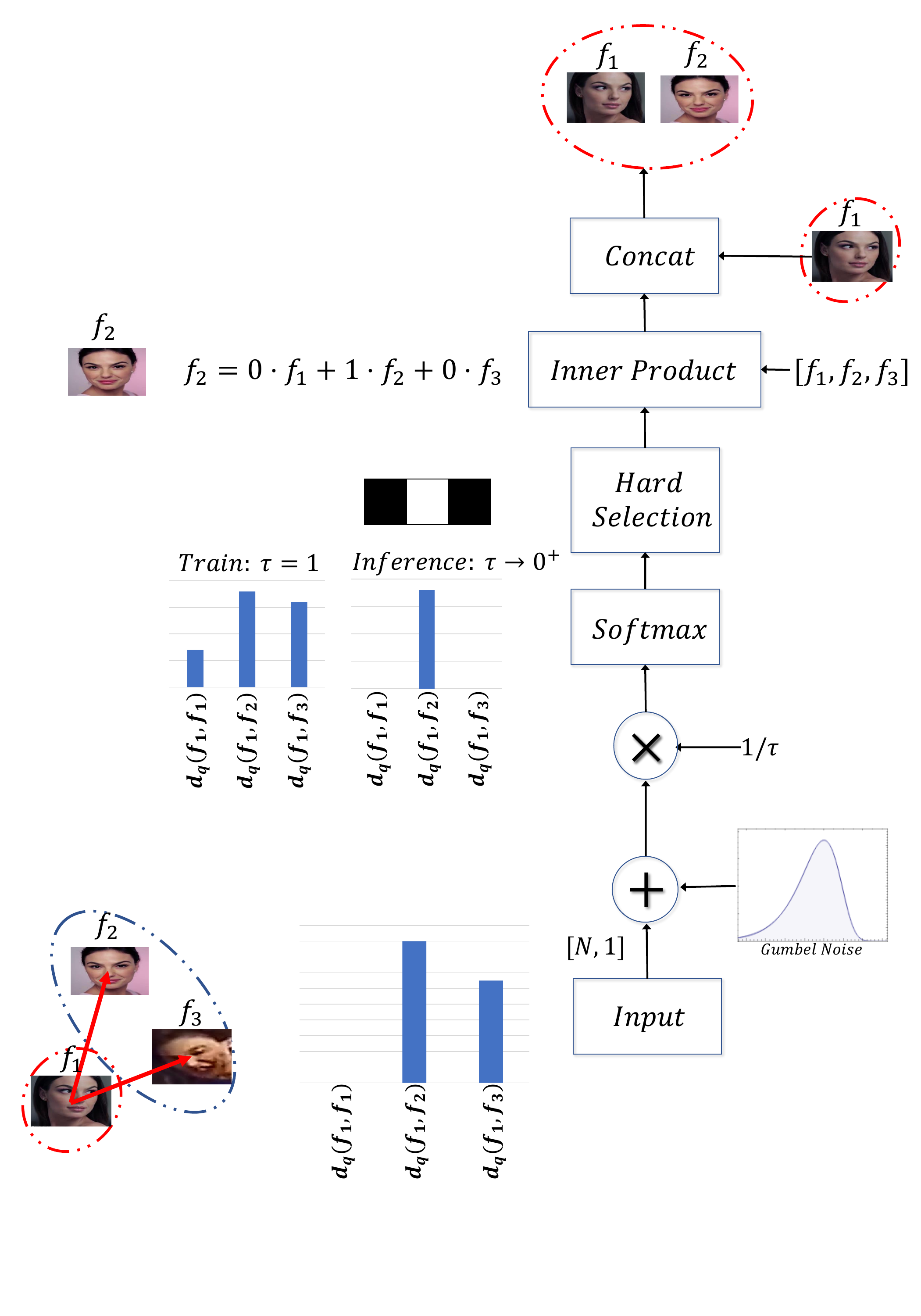}
\centering
\caption{Illustration of a single iteration of the differential core-template selection (lines 19 to 25) in Alg. \ref{alg:core_template_selection} }
\label{fig:soft_selection}
\end{figure}

The algorithm starts by sampling from the Gumbel-Softmax distribution of the template's norms $\bm{F}$ (line 14). The sampling operator returns a 1-hot vector that is used in line 16 to insert the highest quality feature to the Core-Template $\bm{CT}$. In line 18, we calculate the distance from each feature in $\bm{F}$ to the newly selected feature in $\bm{CT}$. In each iteration of the loop starting at line 19 we add one new member to $\bm{CT}$ that is farthest from it. We use the quality aware distance function in line 28 for all the distance calculations.
An example of one iteration in this process is depicted in Figure \ref{fig:soft_selection}. Consider the feature set $\bm{F} = \{\bm{f}_1, \bm{f}_2, \bm{f}_3\}$ and the initial Core-Template $\bm{CT} = \{\bm{f}_1\}$. The quality aware distances from $\bm{F}$ to $\bm{CT}$, serve as input to the selection process.
In the next stage, Gumbel noise is added to each distance value. These values are then multiplied by the temperature parameter $\tau$, and the softmax operation is applied to obtain the final Gumbel-Softmax distribution. A mask is generated to identify the maximal value in the distribution, which is then utilized to determine the next feature to add to $\bm{CT}$.

Importantly, our proposed core template selection process time complexity is $O(NK)$ where $N$ is the template size, and $K$ is the core template size with $K << N$. Assuming $K$ is a small constant (we found $K=3$ optimal in our experiments), we have a selection process with linear complexity. In addition, the core template selection process is permutation invariant, as it depends solely on the geometrical attributes of each feature in $\bm{F}$ and not on their order.

\subsection{Attend}

Our core template selection process performs an important data reduction from unbounded template size $N$ to a small fixed-size core template. This data reduction enables the application of attention mechanisms like multi-head-self-attention \cite{vaswani2017attention}, which have quadratic time complexity and are prohibitively slow for large templates.

We use Multihead Self Attention (MSA) as an encoder to enhance the selected core template representation power and exchange information among its features (Fig. \ref{fig:architecture}).

Inspired by recent developments in detection transformers such as Efficient-DETR \cite{yao2021efficient}, which demonstrate the superiority of leveraging image features as queries instead of relying on learned queries for decoding as in the original transformer \cite{vaswani2017attention} and DETR \cite{carion2020end},
we leverage the core template features as transformer decoder queries that interact with key values taken from the full template features in $\bm{F}$. This allows the core template features to extract additional information from the full template that is missing from the core template.

\textbf{Norm encoding}. It appeals to our intuition that explicit image quality data can aid information extraction in multi-head attention settings, facilitating the suppression of low-quality features and highlighting high-quality features. To incorporate the quality information, we encode the feature norms as a quality proxy, using the standard Sinusidial conversion established in \cite{vaswani2017attention}, and add them to the original features. Our experiments validate the positive impact of Norm encoding on the model’s overall accuracy.

Importantly, our attention process interpolates between template features using self and cross-attention realized with MHA. We avoid the Feed-Forward block in the transformer encoder and decoder blocks, which extrapolates the template features and deteriorates performance.

The complexity of the encoder is $O(K^2)$, which is constant. The complexity of the decoder is $O(NK)$. The total complexity of the attention block is $O(N)$.

\textbf{Aggregate.} To represent the entire template with one feature $\bm{f}$, we combine the $K$ features in the enhanced core template by adding them up and normalizing the result to have a unit length.

\subsection{Training and Loss}
After the aggregation phase, we get a single vector $\bm{f}$ representing the whole template. From this point on, the problem is reduced from SFR to FR, facilitating the use of SoTA FR loss functions. In this work, we choose AdaFace \cite{kim2022adaface} loss for its superior accuracy. AdaFace computes a class margin adaptively based on the image quality of the face approximated by the face feature norm. In our case, the margin is computed by the \textit{template} quality instead of the image (See Fig. \ref{fig:architecture} right-hand side).
Using a simple margin-based loss for template feature fusion facilitates a simplistic implementation compared with elaborate loss terms found in \cite{kim2022adaface} and others.

Large modern FR datasets like WebFace \cite{zhu2021webface260m} contain millions of identities but only a limited number of images per identity (WebFace4M contains about 20 images per identity on average) with no video clips. On the other hand, SFR benchmarks like IJB-B, IJB-C contain rich templates with a diverse combination of still images and video clips for each identity. To bridge this gap, we online augment the training set and simulate IJB-B, IJB-C style templates. To compensate for the lack of video clips, we simulate a video clip by randomly sampling an image and applying various geometrical transformations (translations, rotations, etc.) on it to create multiple "video frames". Each training template comprises a random collection of still photos and simulated video clips with random addition of photometric noise to each image.

\section{Experiments}

To allow for a valid comparison with the current SOTA, we strictly follow the experiments protocol of CAFace \cite{kim2022cluster_aggregate} (NeurIPS'22). Our training dataset is WebFace4M \cite{zhu2021webface260m}, which contains 4.2 million facial images from 205,990 identities. We use CAFaces's pre-trained face recognition model, $E$, an IResnet-101, trained with AdaFace loss on the entire WebFace4M dataset 
. To train the FaceCoresetNet, we use CAFace's subset of 813,482 images from 10,000 identities. As VGG-2 \cite{cao2018vggface2} and MS1MV2 \cite{guo2016ms, deng2019arcface} creators have withheld them due to privacy and other issues, we do not use them in our experiments.
We train FaceCoresetNet for $\bf{2}$ epochs till convergence, compared with CAFace's $\bf{10}$ epochs training schedule.

We test on IJB-B \cite{ijb-b} and IJB-C \cite{ijb-c} datasets as they are intended for template-based face recognition. IJB-B comprises 1845 individuals and includes 11,754 images, 55,025 frames, and 7,011 videos. Each template in the dataset consists of a diverse collection of still images and video frames from various sources. These images and videos were gathered from the Internet and are considered unconstrained, as they exhibit considerable variation in pose, lighting conditions, and image quality. Following CAFace, we focus our experiments on face verification, where IJB-B offers 10,270 genuine template pair comparison and 8,000,000 impostor comparisons, facilitating valid measurements of very small FAR values on the TAR@FAR ROC curve.  IJB-C extends IJB-B by adding 1,661 new subjects, with increased emphasis on occlusion and diversity.

\begin{table}[]
\resizebox{\linewidth}{!}{%
\begin{tabular}{@{}cccccc@{}}
\toprule
\# & \begin{tabular}[c]{@{}c@{}}Differential \\ Core-Template \\ Selection\end{tabular} & \multicolumn{1}{l}{Self Att} & \multicolumn{1}{l}{Cross Att} & \multicolumn{1}{l}{\begin{tabular}[c]{@{}l@{}}Norm\\ Encoding\end{tabular}} & \multicolumn{1}{l}{\begin{tabular}[c]{@{}l@{}}IJB-B:TAR@\\ FAR=1e-6\end{tabular}} \\ \midrule
1  & $\times$                                                                                  & $\times$                            & $\times$                             & $\times$                                                                           &    38.45                                                                         \\
2  & \checkmark                                                                                  & $\times$                            & $\times$                             & $\times$                                                                           &     48.01                                                                        \\
3  & \checkmark                                                                                  & \checkmark                            & $\times$                             & $\times$                                                                           &    49.24                                                                         \\
4  & \checkmark                                                                                  & \checkmark                            & \checkmark                             & $\times$                                                                           &   51.71                                                                          \\
5  & \checkmark                                                                                  & \checkmark                            & \checkmark                             & \checkmark                                                                           &    52.56                                                                         \\ \bottomrule
\end{tabular}}
\caption{FaceCoresetNet Ablations}
\label{table:ablations}
\end{table}

\begin{table*}[]
\resizebox{\textwidth}{!}{%
\begin{tabular}{@{}lllcccccc@{}}
\toprule
\multicolumn{1}{c}{}                         & \multicolumn{1}{c}{}                          &                                                                                      & \multicolumn{3}{c}{IJB-B 1:1 Verification TAR}                                                                                       & \multicolumn{3}{c}{IJB-C 1:1 Verification TAR}                                                                        \\ \cmidrule(l){4-9} 
\multicolumn{1}{c}{\multirow{-2}{*}{Method}} & \multirow{-2}{*}{\begin{tabular}[c]{@{}l@{}} Complexity \\($N$: Template Size)\end{tabular}} & \multicolumn{1}{l}{FAR=1e-4}          & \multicolumn{1}{l}{FAR=1e-5}          & \multicolumn{1}{l}{FAR=1e-6}                         & \multicolumn{1}{l}{FAR=1e-4}          & \multicolumn{1}{l}{FAR=1e-5}          & \multicolumn{1}{l}{FAR=1e-6}          \\ \midrule
AdaFace \cite{kim2022adaface}                                      & NA                                                                                   & 94.84                                 & 90.86                                 & 38.45                                                & 96.42                                 & 94.47                                 & 87.90                                 \\

RSA \cite{liu2019permutation}                                                  & $O(N^2)$& 95.00 & {\color[HTML]{FE0000} 91.22} & - & 96.49 & 94.58 & - \\
SD + VBA \cite{wang2022set}                                     & $O(N^2)$                                                                             & {\color[HTML]{3166FF} 95.38}                                 & 90.89          & \cellcolor[HTML]{FFFFFF}{\color[HTML]{3166FF} 48.96} & {\color[HTML]{FE0000} 96.65}                                 & {\color[HTML]{FE0000} 94.85}                                 & {\color[HTML]{FE0000} 90.02}          \\
CAFace \cite{kim2022cluster_aggregate}                                                                            & $O(N^2)$                                                                             & {\color[HTML]{34FF34} \textbf{95.78}} & {\color[HTML]{34FF34} \textbf{92.78}} & {\color[HTML]{FE0000} 47.21}                         & {\color[HTML]{34FF34} \textbf{97.30}} & {\color[HTML]{34FF34} \textbf{95.96}} & {\color[HTML]{3166FF} 90.56}          \\
\textbf{FaceCoresetNet}                                                           & \bm{$O(N)$}                                                                               & {\color[HTML]{FE0000} 95.28 }                                & {\color[HTML]{3166FF} 91.44}          & {\color[HTML]{34FF34} \textbf{52.56}}                & {\color[HTML]{3166FF} 96.96}          & {\color[HTML]{3166FF} 95.21}          & {\color[HTML]{34FF34} \textbf{91.12}} \\ \bottomrule
\end{tabular}}
\caption{A performance comparison of recent methods on IJB-B \cite{ijb-b} and IJB-C \cite{ijb-c} datasets. The results are color-coded, with green representing the best, blue the second best, and red the third-best outcomes. FaceCoresetNet achieves the best results on TPR@FPR=1e-6 for both IJB-B and IJB-C while being the most efficient.}
\label{table:comparison_with_sota}
\end{table*}





\subsection{Ablation and Analysis}

To demonstrate the effectiveness of each component in FaceCoresetnNet, we conduct a series of experiments by ablating individual components. Initially, we turned off all components, resulting in a configuration where the template feature was simply the average feature computed from all the templates' features. The obtained TAR@FPR=1e-6 for the IJB-B dataset was 38.45 (Table \ref{table:ablations}).
Subsequently, we enabled the differential core-template selection. We computed the final template descriptor by averaging the pooled features within the core template. Surprisingly, by effectively selecting 3 features, the accuracy increased significantly to 48.01. We hypothesize that our core template's high quality and diversity reduce the amount of noise present in the original template, which may explain the observed improvement in accuracy.
Furthermore, we integrated self-attention into the core-template features, boosting accuracy to 49.24. Additionally, by employing cross-attention between the core and full templates, the accuracy increased to 51.71. Finally, with the incorporation of Norm encoding, the accuracy reached 52.56.

\begin{table}[]
\small
\centering
\begin{tabular}{@{}cc@{}}
\toprule
\multicolumn{1}{l}{$\gamma$ policy} & \begin{tabular}[c]{@{}c@{}}IJB-C: \\ TAR@FAR=1e-6\end{tabular} \\ \midrule
Fixed 0 (Diversity priority)                                      &         90.84                                                      \\
Fixed 10 (Quality priority)                                      &              89.69                                                 \\
Trained                                      &   \bf{91.12}               
\\ \bottomrule
\end{tabular}
\caption{Training $\gamma$ for optimal balance between quality and diversity yields best results, compared with fixed $\gamma$ values}
\label{table:gamma_policy}
\end{table}


\textbf{Effect of Core-Template size}. To determine the optimal core-template size, we searched a range of sizes from 1 to 6. The outcomes of this search are summarized in Table 1 of the supplementary materials. Notably, the ideal coreset size was identified as 3, and this optimal value remained constant throughout all our subsequent experiments.

\textbf{$\gamma$ computation policy}. Our model optimizes the $\gamma$ value in the quality-aware distance. To showcase the effectiveness of this optimization, we conducted 3 experiments. One setting $\gamma$ to the constant 10 value, prioritizing quality over diversity. Conversely, we set $\gamma$ to 0, which reduces the quality aware distance to the vanilla cosine distance, prioritizing diversity over quality. In the last experiment, we optimize $\gamma$ end-to-end. The results in Table \ref{table:gamma_policy} demonstrates the effectiveness of our approach.

\textbf{Comparison with SoTA methods}. In Table. \ref{table:comparison_with_sota}, we compare our results with the state-of-the-art (SOTA) in face verification, focusing on the IJB-B and IJB-C datasets. Our method achieves the highest accuracy in the low false acceptance rate regime, TAR@FAR=10e-6, for both IJB-B and IJB-C, outperforming others significantly. Additionally, we obtain the second-best results for TAR@FAR=10e-5 while maintaining the advantage of utilizing the lowest \textbf{linear} time compute.

\begin{figure}

\includegraphics[clip, trim=2.5cm 0.5cm 3cm 1.5cm,width=0.8\linewidth]{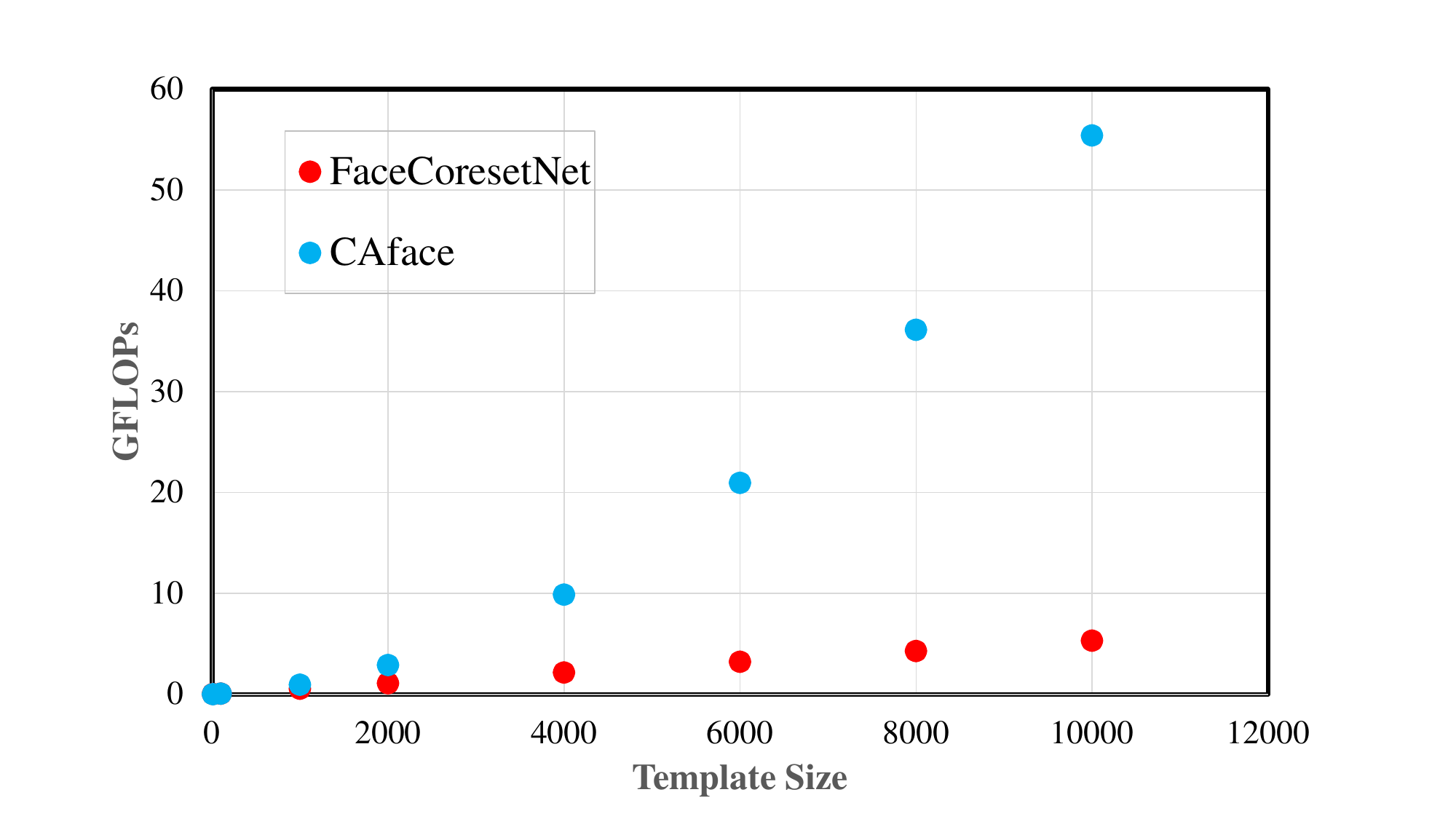}
\centering
\caption{Computation complexity comparison with SoTA. FaceCoresetNet is linear in the template size, while CAFace is quadratic.}
\label{fig:flops}
\end{figure}

\textbf{Computation Complexity}. To validate our algorithm's claimed linear time complexity, we employ \cite{fvcore} to calculate the FLOPS in our FaceCoresetNet and compare it against CAFace \cite{kim2022cluster_aggregate} using the authors' code. The FLOPS measurements for different template sizes are illustrated in Fig. \ref{fig:flops}. The results demonstrate that our algorithm indeed operates in linear time, while CAFace exhibits quadratic time complexity.

\section{Conclusion}
In this work, we employ a method to fuse face template features by selecting a small, fixed-size core-template that balances quality and diversity. This selection uses a differentiable farthest-point sampling with a quality-aware distance metric, realized by sampling from the Gumbel-Softmax distribution of quality-aware distances.

The selected core template features are further enhanced using MHA self-attention. These enhanced features are then used as queries for MHA cross-attention between the core template and the entire template, enabling the extraction of additional information that may be missing from the core template.

Our proposed model achieves state-of-the-art accuracy on IJB-B and IJB-C datasets, all while maintaining reduced linear time complexity and simple design.
FaceCoresetNet is showcased within the context of face set recognition, yet its applicability may extend to other domains requiring efficient feature fusion based on a differentiable selection with a learned metric. We hope to see an extension of this work to other domains in future research.

\bibliography{FaceCoresetNet}

\begin{thebibliography}{41}
\providecommand{\natexlab}[1]{#1}

\bibitem[{Agarwal et~al.(2005)Agarwal, Har-Peled, Varadarajan
  et~al.}]{agarwal2005geometric}
Agarwal, P.~K.; Har-Peled, S.; Varadarajan, K.~R.; et~al. 2005.
\newblock Geometric approximation via coresets.
\newblock \emph{Combinatorial and computational geometry}, 52(1): 1--30.

\bibitem[{Campbell and Broderick(2018)}]{campbell2018bayesian_coreset}
Campbell, T.; and Broderick, T. 2018.
\newblock Bayesian coreset construction via greedy iterative geodesic ascent.
\newblock In \emph{Proceedings of the International Conference on Machine
  Learning (ICML)}, 698--706. PMLR.

\bibitem[{Cao et~al.(2018)Cao, Shen, Xie, Parkhi, and
  Zisserman}]{cao2018vggface2}
Cao, Q.; Shen, L.; Xie, W.; Parkhi, O.~M.; and Zisserman, A. 2018.
\newblock Vggface2: A dataset for recognising faces across pose and age.
\newblock In \emph{International Conference on Automatic Face and Gesture
  Recognition (FGR)}, 67--74. IEEE.

\bibitem[{Carion et~al.(2020)Carion, Massa, Synnaeve, Usunier, Kirillov, and
  Zagoruyko}]{carion2020end}
Carion, N.; Massa, F.; Synnaeve, G.; Usunier, N.; Kirillov, A.; and Zagoruyko,
  S. 2020.
\newblock End-to-end object detection with transformers.
\newblock In \emph{Proceedings of the European Conference on Computer Vision
  (ECCV)}, 213--229. Springer.

\bibitem[{Chen et~al.(2015)Chen, Ranjan, Kumar, Chen, Patel, and
  Chellappa}]{chen2015end_avgpool}
Chen, J.-C.; Ranjan, R.; Kumar, A.; Chen, C.-H.; Patel, V.~M.; and Chellappa,
  R. 2015.
\newblock An end-to-end system for unconstrained face verification with deep
  convolutional neural networks.
\newblock In \emph{Proceedings of the IEEE International Conference on Computer
  Vision Workshops (ICCVW)}, 118--126.

\bibitem[{Deng et~al.(2019)Deng, Guo, Xue, and Zafeiriou}]{deng2019arcface}
Deng, J.; Guo, J.; Xue, N.; and Zafeiriou, S. 2019.
\newblock {ArcFace}: Additive angular margin loss for deep face recognition.
\newblock In \emph{Proceedings of the IEEE/CVF Conference on Computer Vision
  and Pattern Recognition (CVPR)}, 4690--4699.

\bibitem[{Dosovitskiy et~al.(2021)Dosovitskiy, Beyer, Kolesnikov, Weissenborn,
  Zhai, Unterthiner, Dehghani, Minderer, Heigold, Gelly, Uszkoreit, and
  Houlsby}]{dosovitskiy2020vit}
Dosovitskiy, A.; Beyer, L.; Kolesnikov, A.; Weissenborn, D.; Zhai, X.;
  Unterthiner, T.; Dehghani, M.; Minderer, M.; Heigold, G.; Gelly, S.;
  Uszkoreit, J.; and Houlsby, N. 2021.
\newblock An Image is Worth 16x16 Words: Transformers for Image Recognition at
  Scale.
\newblock \emph{Proceedings of the International Conference on Learning
  Representations (ICLR)}.

\bibitem[{Eldar et~al.(1997)Eldar, Lindenbaum, Porat, and
  Zeevi}]{eldar1997farthest}
Eldar, Y.; Lindenbaum, M.; Porat, M.; and Zeevi, Y.~Y. 1997.
\newblock The farthest point strategy for progressive image sampling.
\newblock \emph{IEEE Transactions on Image Processing}, 6(9): 1305--1315.

\bibitem[{Feldman, Faulkner, and Krause(2011)}]{feldman2011scalable_coreset}
Feldman, D.; Faulkner, M.; and Krause, A. 2011.
\newblock Scalable Training of Mixture Models via Coresets.
\newblock In Shawe-Taylor, J.; Zemel, R.; Bartlett, P.; Pereira, F.; and
  Weinberger, K., eds., \emph{Advances in Neural Information Processing Systems
  (NIPS)}, volume~24. Curran Associates, Inc.

\bibitem[{fvcore(2023)}]{fvcore}
fvcore. 2023.
\newblock fvcore.
\newblock \url{https://github.com/facebookresearch/fvcore}.

\bibitem[{Gong, Shi, and Jain(2019)}]{gong2019low}
Gong, S.; Shi, Y.; and Jain, A. 2019.
\newblock Low quality video face recognition: Multi-mode aggregation recurrent
  network (MARN).
\newblock In \emph{IEEE Conference on Computer Vision and Pattern Recognition
  Workshops (CVPRW)}, 0--0.

\bibitem[{Gong et~al.(2019)Gong, Shi, Kalka, and Jain}]{gong2019video}
Gong, S.; Shi, Y.; Kalka, N.~D.; and Jain, A.~K. 2019.
\newblock Video face recognition: Component-wise feature aggregation network
  (c-fan).
\newblock In \emph{International Conference on Biometrics (ICB)}, 1--8. IEEE.

\bibitem[{Graves and Graves(2012)}]{graves2012long}
Graves, A.; and Graves, A. 2012.
\newblock Long short-term memory.
\newblock \emph{Supervised sequence labelling with recurrent neural networks},
  37--45.

\bibitem[{Guo et~al.(2016)Guo, Zhang, Hu, He, and Gao}]{guo2016ms}
Guo, Y.; Zhang, L.; Hu, Y.; He, X.; and Gao, J. 2016.
\newblock Ms-celeb-1m: A dataset and benchmark for large-scale face
  recognition.
\newblock In \emph{Proceedings of the European Conference on Computer Vision
  (ECCV)}, 87--102. Springer.

\bibitem[{Har-Peled and Kushal(2005)}]{har2005smaller}
Har-Peled, S.; and Kushal, A. 2005.
\newblock Smaller coresets for k-median and k-means clustering.
\newblock In \emph{Proceedings of the twenty-first annual symposium on
  Computational geometry}, 126--134.

\bibitem[{Jang, Gu, and Poole(2016)}]{jang2016categorical}
Jang, E.; Gu, S.; and Poole, B. 2016.
\newblock Categorical reparameterization with gumbel-softmax.
\newblock \emph{arXiv preprint arXiv:1611.01144}.

\bibitem[{Kalka et~al.(2018)Kalka, Maze, Duncan, O’Connor, Elliott, Hebert,
  Bryan, and Jain}]{kalka2018ijb}
Kalka, N.~D.; Maze, B.; Duncan, J.~A.; O’Connor, K.; Elliott, S.; Hebert, K.;
  Bryan, J.; and Jain, A.~K. 2018.
\newblock Ijb--s: Iarpa janus surveillance video benchmark.
\newblock In \emph{2018 IEEE 9th international conference on biometrics theory,
  applications and systems (BTAS)}, 1--9. IEEE.

\bibitem[{Kim, Jain, and Liu(2022)}]{kim2022adaface}
Kim, M.; Jain, A.~K.; and Liu, X. 2022.
\newblock {AdaFace}: Quality adaptive margin for face recognition.
\newblock In \emph{Proceedings of the IEEE/CVF Conference on Computer Vision
  and Pattern Recognition (CVPR)}, 18750--18759.

\bibitem[{Kim et~al.(2022)Kim, Liu, Jain, and Liu}]{kim2022cluster_aggregate}
Kim, M.; Liu, F.; Jain, A.; and Liu, X. 2022.
\newblock Cluster and Aggregate: Face Recognition with Large Probe Set.
\newblock In \emph{Advances in Neural Information Processing Systems (NIPS)}.

\bibitem[{Kim and Shin(2022)}]{kim2022defense_coreset}
Kim, Y.; and Shin, B. 2022.
\newblock In Defense of Core-set: A Density-aware Core-set Selection for Active
  Learning.
\newblock In \emph{Proceedings of the 28th ACM SIGKDD Conference on Knowledge
  Discovery and Data Mining}, 804--812.

\bibitem[{Liu et~al.(2019)Liu, Guo, Li, Kong, Jia, You, and
  Kumar}]{liu2019permutation}
Liu, X.; Guo, Z.; Li, S.; Kong, L.; Jia, P.; You, J.; and Kumar, B. 2019.
\newblock Permutation-invariant feature restructuring for correlation-aware
  image set-based recognition.
\newblock In \emph{Proceedings of the IEEE International Conference on Computer
  Vision (ICCV)}, 4986--4996.

\bibitem[{Liu et~al.(2018)Liu, Kumar, Yang, Tang, and You}]{liu2018dependency}
Liu, X.; Kumar, B.; Yang, C.; Tang, Q.; and You, J. 2018.
\newblock Dependency-aware attention control for unconstrained face recognition
  with image sets.
\newblock In \emph{Proceedings of the European Conference on Computer Vision
  (ECCV)}, 548--565.

\bibitem[{Liu, Yan, and Ouyang(2017)}]{liu2017quality}
Liu, Y.; Yan, J.; and Ouyang, W. 2017.
\newblock Quality aware network for set to set recognition.
\newblock In \emph{Proceedings of the IEEE/CVF Conference on Computer Vision
  and Pattern Recognition (CVPR)}, 5790--5799.

\bibitem[{Maze et~al.(2018)Maze, Adams, Duncan, Kalka, Miller, Otto, Jain,
  Niggel, Anderson, Cheney, and Grother}]{ijb-c}
Maze, B.; Adams, J.; Duncan, J.~A.; Kalka, N.; Miller, T.; Otto, C.; Jain,
  A.~K.; Niggel, W.~T.; Anderson, J.; Cheney, J.; and Grother, P. 2018.
\newblock IARPA Janus Benchmark - C: Face Dataset and Protocol.
\newblock In \emph{International Conference on Biometrics (ICB)}, 158--165.

\bibitem[{Meng et~al.(2021)Meng, Zhao, Huang, and Zhou}]{meng2021magface}
Meng, Q.; Zhao, S.; Huang, Z.; and Zhou, F. 2021.
\newblock Magface: A universal representation for face recognition and quality
  assessment.
\newblock In \emph{Proceedings of the IEEE/CVF Conference on Computer Vision
  and Pattern Recognition (CVPR)}, 14225--14234.

\bibitem[{Mirzasoleiman, Bilmes, and
  Leskovec(2020)}]{mirzasoleiman2020coresets}
Mirzasoleiman, B.; Bilmes, J.; and Leskovec, J. 2020.
\newblock Coresets for data-efficient training of machine learning models.
\newblock In \emph{International Conference on Machine Learning}, 6950--6960.
  PMLR.

\bibitem[{Mussay et~al.(2021)Mussay, Feldman, Zhou, Braverman, and
  Osadchy}]{mussay2021datacoreset}
Mussay, B.; Feldman, D.; Zhou, S.; Braverman, V.; and Osadchy, M. 2021.
\newblock Data-independent structured pruning of neural networks via coresets.
\newblock \emph{IEEE Transactions on Neural Networks and Learning Systems},
  33(12): 7829--7841.

\bibitem[{Parkhi, Vedaldi, and Zisserman(2015)}]{parkhi2015deep_avgpool}
Parkhi, O.~M.; Vedaldi, A.; and Zisserman, A. 2015.
\newblock Deep face recognition.

\bibitem[{Pooladzandi, Davini, and
  Mirzasoleiman(2022)}]{pooladzandi2022adaptive_coreset}
Pooladzandi, O.; Davini, D.; and Mirzasoleiman, B. 2022.
\newblock Adaptive second order coresets for data-efficient machine learning.
\newblock In \emph{Proceedings of the International Conference on Machine
  Learning (ICML)}, 17848--17869. PMLR.

\bibitem[{Roth et~al.(2020)Roth, Milbich, Sinha, Gupta, Ommer, and
  Cohen}]{roth2020revisiting_coreset}
Roth, K.; Milbich, T.; Sinha, S.; Gupta, P.; Ommer, B.; and Cohen, J.~P. 2020.
\newblock Revisiting training strategies and generalization performance in deep
  metric learning.
\newblock In \emph{Proceedings of the International Conference on Machine
  Learning (ICML)}, 8242--8252. PMLR.

\bibitem[{Sinha et~al.(2020)Sinha, Zhang, Goyal, Bengio, Larochelle, and
  Odena}]{sinha2020small_coreset}
Sinha, S.; Zhang, H.; Goyal, A.; Bengio, Y.; Larochelle, H.; and Odena, A.
  2020.
\newblock Small-gan: Speeding up gan training using core-sets.
\newblock In \emph{Proceedings of the International Conference on Machine
  Learning (ICML)}, 9005--9015. PMLR.

\bibitem[{Vaswani et~al.(2017)Vaswani, Shazeer, Parmar, Uszkoreit, Jones,
  Gomez, Kaiser, and Polosukhin}]{vaswani2017attention}
Vaswani, A.; Shazeer, N.; Parmar, N.; Uszkoreit, J.; Jones, L.; Gomez, A.~N.;
  Kaiser, {\L}.; and Polosukhin, I. 2017.
\newblock Attention is all you need.
\newblock \emph{Advances in Neural Information Processing Systems (NIPS)}, 30.

\bibitem[{Wang, Zhao, and Wu(2022)}]{wang2022set}
Wang, J.; Zhao, Z.; and Wu, F. 2022.
\newblock Set-Based Face Recognition Beyond Disentanglement: Burstiness
  Suppression With Variance Vocabulary.
\newblock In \emph{Proceedings of the 30th ACM International Conference on
  Multimedia}, 6125--6135.

\bibitem[{Wang et~al.(2018)Wang, Girshick, Gupta, and He}]{wang2018non}
Wang, X.; Girshick, R.; Gupta, A.; and He, K. 2018.
\newblock Non-local neural networks.
\newblock In \emph{Proceedings of the IEEE/CVF Conference on Computer Vision
  and Pattern Recognition (CVPR)}, 7794--7803.

\bibitem[{Whitelam et~al.(2017)Whitelam, Taborsky, Blanton, Maze, Adams,
  Miller, Kalka, Jain, Duncan, Allen, Cheney, and Grother}]{ijb-b}
Whitelam, C.; Taborsky, E.; Blanton, A.; Maze, B.; Adams, J.; Miller, T.;
  Kalka, N.; Jain, A.~K.; Duncan, J.~A.; Allen, K.; Cheney, J.; and Grother, P.
  2017.
\newblock IARPA Janus Benchmark-B Face Dataset.
\newblock In \emph{IEEE Conference on Computer Vision and Pattern Recognition
  Workshops (CVPRW)}, 592--600.

\bibitem[{Xie and Zisserman(2018)}]{xie2018multicolumn}
Xie, W.; and Zisserman, A. 2018.
\newblock Multicolumn networks for face recognition.
\newblock \emph{arXiv preprint arXiv:1807.09192}.

\bibitem[{Yang et~al.(2019)Yang, Zhang, Ni, Li, Liu, Zhou, and
  Tian}]{yang2019modeling}
Yang, J.; Zhang, Q.; Ni, B.; Li, L.; Liu, J.; Zhou, M.; and Tian, Q. 2019.
\newblock Modeling point clouds with self-attention and gumbel subset sampling.
\newblock In \emph{Proceedings of the IEEE/CVF Conference on Computer Vision
  and Pattern Recognition (CVPR)}, 3323--3332.

\bibitem[{Yao et~al.(2021)Yao, Ai, Li, and Zhang}]{yao2021efficient}
Yao, Z.; Ai, J.; Li, B.; and Zhang, C. 2021.
\newblock Efficient detr: improving end-to-end object detector with dense
  prior.
\newblock \emph{arXiv preprint arXiv:2104.01318}.

\bibitem[{Zhang et~al.(2022)Zhang, Li, Liu, Zhang, Su, Zhu, Ni, and
  Shum}]{zhang2022dino}
Zhang, H.; Li, F.; Liu, S.; Zhang, L.; Su, H.; Zhu, J.; Ni, L.; and Shum, H.
  2022.
\newblock Dino: Detr with improved denoising anchor boxes for end-to-end object
  detection.
\newblock In \emph{Proceedings of the International Conference on Learning
  Representations (ICLR)}.

\bibitem[{Zheng et~al.(2021)Zheng, Lu, Zhao, Zhu, Luo, Wang, Fu, Feng, Xiang,
  Torr et~al.}]{zheng2021rethinking}
Zheng, S.; Lu, J.; Zhao, H.; Zhu, X.; Luo, Z.; Wang, Y.; Fu, Y.; Feng, J.;
  Xiang, T.; Torr, P.~H.; et~al. 2021.
\newblock Rethinking semantic segmentation from a sequence-to-sequence
  perspective with transformers.
\newblock In \emph{Proceedings of the IEEE/CVF Conference on Computer Vision
  and Pattern Recognition (CVPR)}, 6881--6890.

\bibitem[{Zhu et~al.(2021)Zhu, Huang, Deng, Ye, Huang, Chen, Zhu, Yang, Lu, Du
  et~al.}]{zhu2021webface260m}
Zhu, Z.; Huang, G.; Deng, J.; Ye, Y.; Huang, J.; Chen, X.; Zhu, J.; Yang, T.;
  Lu, J.; Du, D.; et~al. 2021.
\newblock Webface260m: A benchmark unveiling the power of million-scale deep
  face recognition.
\newblock In \emph{Proceedings of the IEEE/CVF Conference on Computer Vision
  and Pattern Recognition (CVPR)}, 10492--10502.

\end{thebibliography}

\end{document}


\maketitle

\subsection{Implementation Detail}

To train FaceCoresetNet, we utilize a pre-trained IResenet101 backbone from WebFace4M \cite{zhu2021webface260m} with AdaFace \cite{kim2022adaface} loss. The training set comprises 10,000 identities from WebFace4M, identical to \cite{kim2022cluster_aggregate}.

To simulate templates resembling the IJB \cite{ijb-b} style, we construct batches with template size N, randomly chosen between Nmin=1 and Nmax=20. Once the template size N is determined, we compose an N-size template by randomly selecting single frames and simulated video clips. These simulated video clips are generated by augmenting a 'seed' frame with random geometric transformations to create clips of various sizes.
The learning rate is set to $1e - 4$, weight decay is set $1e - 3$. We use the following AdaFace loss parameters: $s=48$, $m=0.8$, $h=0.333$.

Due to GPU memory constraints, we set the batch size to 20 and perform training on two Nvidia RTX 2080ti GPUs. Our model is trained for 2 epochs, compared to the 10 epochs used in CAFace.



\subsection{Finding the optimal Core-Template size}

\begin{figure}

\includegraphics[clip, trim=2cm 18.5cm 7cm 2cm,width=1\linewidth]{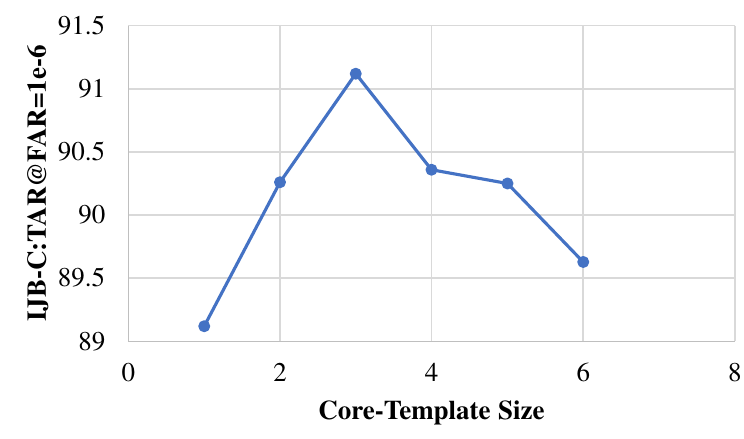}
\centering
\caption{Core template size vs. Accuracy. The optimal size is 3}
\label{fig:coreset_size_search}
\end{figure}

To find the optimal Core-Template size, we trained 6 models with 
varied Core-Template size while keeping other hyper-parameters fixed to identify the best size. Results (Fig. \ref{fig:coreset_size_search}) show peak performance at $K=3$; beyond that, accuracy declined. We  assume larger Core-Template sizes introduce more low-quality images, impairing the template descriptor discrimination power.



\subsection{How the Core-Template size is related to the optimized $\gamma$ value}
\begin{figure}
\includegraphics[clip, trim=2cm 18.5cm 7cm 2cm,width=1\linewidth]{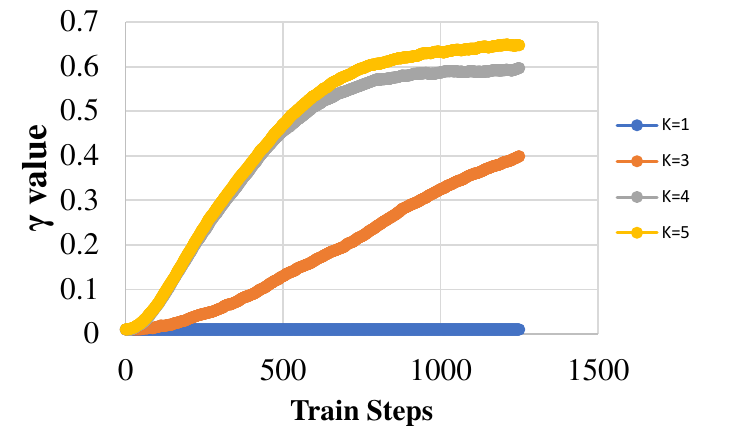}
\centering
\caption{$\gamma$ value per step during training for different Core-Template sizes. The larger the core-template, the higher the final optimized $\gamma$ value. When $K = 1$, $\gamma$ plays no role in the core-template selection, hence it remains constant (blue curve).}
\label{fig:coreset_size_vs_gamma}
\end{figure}

The paper discusses the role of the $\gamma$ parameter within the quality-aware metric, which strikes a balance between diversity and quality in the differentiable Core-Template selection method. As the Core-Template size increases, diversity naturally grows. Therefore, we predict the model will assign more weight to quality (higher $\gamma$ values) when the Core-Template size is larger. Conversely, diversity becomes more crucial with a smaller Core-Template, suggesting lower $\gamma$ values. To validate this notion, we trained four models with varying Core-Template sizes. The outcomes are illustrated in Figure \ref{fig:coreset_size_vs_gamma}. Notably, for a Core-Template featuring just one element ($K=1$), where the highest norm feature is selected without utilizing the quality-aware metric, $\gamma$ remains constant (gray curve). Upon increasing $K$, the training session consistently elevates the $\gamma$ values, supporting our hypothesis.

\subsection{The efficacy of optimizing $\gamma$} 
\begin{figure}
\includegraphics[clip, trim=2cm 18.5cm 7cm 2cm,width=1\linewidth]{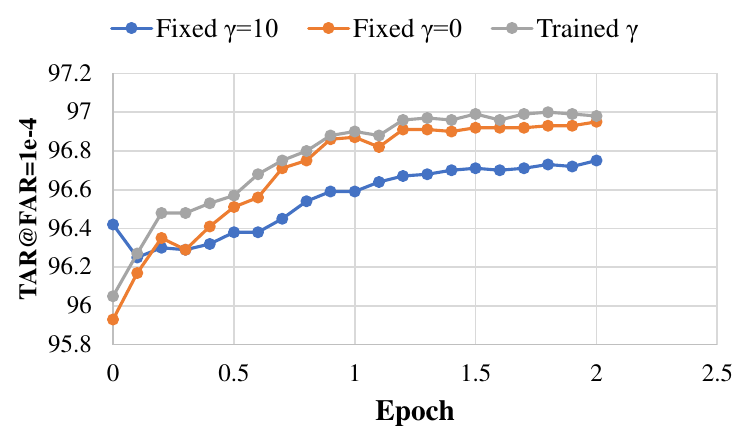}
\centering
\caption{Contrast among various $\gamma$ computation strategies: The optimal accuracy is achieved by computing $\gamma$ through end-to-end optimization (illustrated by the gray curve). This contrasts with fixing $\gamma$ at a constant value of 10, emphasizing  quality in Core-Template selection, or setting $\gamma$ to 0, which transforms the distance metric into cosine distance, emphasizing diversity instead. The accuracy is measured on IJB-C}
\label{fig:gamma_comparison}
\end{figure}

We experiment with three models to show that optimizing the parameter $\gamma$ (as explained in the paper) is better than keeping it fixed. In the first model, we set $\gamma$ to 0, focusing solely on diversity and neglecting quality. In the second model $\gamma$ set to 10, prioritizing quality over diversity. In the third model, we optimize $\gamma$ through backpropagation during end-to-end training. The results in Fig. \ref{fig:gamma_comparison} show the best accuracy when $\gamma$ is trained end-to-end, supporting our claims.

\subsection{Qualitative Experiments}

\begin{figure*}
\includegraphics[clip, trim=1cm 8cm 4cm 3cm,width=1\linewidth]{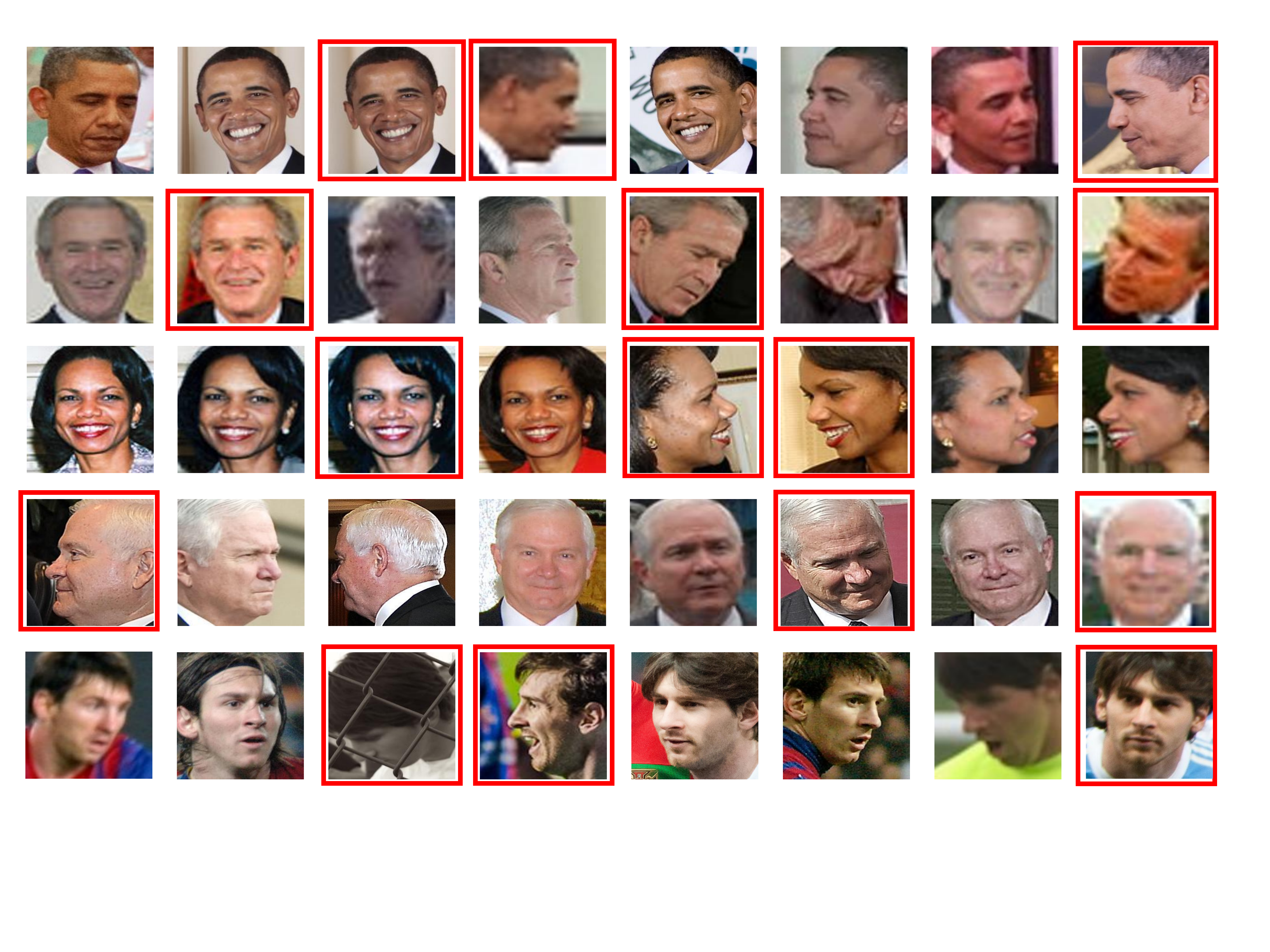}
\centering
\caption{Qualitative differentiable core-template selection experiments: Each row depicts a template with 8 images of a different individual from IJB-C. The selected core template is highlighted by the 3 red squares in each row. 
Notably, the selected core template exhibits high-quality images with diverse appearance.
}
\label{fig:qualitaive_exp}
\end{figure*}

To visually demonstrate our Differentiable Core-Template selection approach, we apply it on several templates from IJB-C, as shown in Figure \ref{fig:qualitaive_exp}. For this process, we utilized a Core-Template containing 3 features ($K=3$). Red rectangles identify the images chosen for the Core-Template. These instances illustrate how the model picks images with diverse appearances and high quality.

\bigskip

\bibliography{FaceCoresetNet}